# OSAD: Open-Set Aircraft Detection in SAR Images

Xiayang Xiao, *Zhuoxuan Li, Student Member IEEE*, Haipeng Wang, Senior *Member, IEEE*

*Abstract*—Current mainstream SAR image object detection methods still lack robustness when dealing with unknown objects in open environments. Open-set detection aims to enable detectors trained on a closed set to detect all known objects and identify unknown objects in open-set environments. The key challenges are how to improve the generalization to potential unknown objects and reduce the empirical classification risk of known categories under strong supervision. To address these challenges, a novel open-set aircraft detector for SAR images is proposed, named Open-Set Aircraft Detection (OSAD), which is equipped with three dedicated components: global context modeling (GCM), location quality-driven pseudo labeling generation (LPG), and prototype contrastive learning (PCL). GCM effectively enhances the network's representation of objects by attention maps which is formed through the capture of long sequential positional relationships. LPG leverages clues about object positions and shapes to optimize localization quality, avoiding overfitting to known category information and enhancing generalization to potential unknown objects. PCL employs prototype-based contrastive encoding loss to promote instance-level intra-class compactness and inter-class variance, aiming to minimize the overlap between known and unknown distributions and reduce the empirical classification risk of known categories. Extensive experiments have demonstrated that the proposed method can effectively detect unknown objects and exhibit competitive performance without compromising closed-set performance. The highest absolute gain which ranges from 0 to 18.36% can be achieved on the average precision of unknown objects.

## I. INTRODUCTION

DUE to the continuous progress in deep learning and convolutional neural network (CNN), visual object detection has witnessed significant advancements in recent years [1]-[3]. Concurrently, deep learning techniques have been introduced into the remote sensing domain, especially in tasks like synthetic aperture radar (SAR) object detection and recognition, where they play a pivotal role [4]-[8]. However, most contemporary methods are built on a strong assumption, referred to as closed-set modeling, which necessitates knowledge of all categories to be detected during the training phase. This modeling approach implies that the system can only be modeled based on observed categories. Under this assumption, object detectors might exhibit inaccuracies when dealing with objects from unknown categories, either misclassifying them as background objects or incorrectly classifying them into established categories.

As per the findings in [9], when conducting tests using samples from sources distinct from the training set, the performance of the optimal system for object classification and recognition is significantly compromised. More specifically, owing to constraints in the training dataset, only a limited subset of categories can be included. This leads to a lack of resilience in the majority of detection and recognition methods when confronted with unknown objects. In typical CNN, as depicted in Equation (1), the confidence score of the classification layer is conventionally employed for identifying unknown categories [10]. In essence, the determination of classifying an object as an unknown category relies on the confidence score generated by the network's output.

$$\text{conf}(x) = \max_y p(y|x) = \max_y \frac{e^{z_y}}{\sum_{i=1}^{C_k} e^{z_i}} \quad (1)$$

Where $z_i$ represents the logit value for the $i$-th class, if the confidence score conf(x) for a sample x surpasses the threshold T, it is classified as a known category; otherwise, it is labeled as unknown. During the training process, the posterior probabilities of the $C_k$ known categories are normalized to yield a sum of 1, which frequently results in unknown objects being incorrectly assigned to existing categories with high confidence. These errors can be broadly classified into two cases, as shown in Fig.1: 1) false negatives, where targets are erroneously classified as other categories or as background; and 2) false positives, where background or unknown objects are mistakenly recognized as existing targets.

This issue can be attributed to several reasons:1) Algorithmic design: In closed-set training, it is common practice to normalize the probabilities of known categories to 1 through the SoftMax layer, leaving no probability space for unknown categories; 2) Feature level: The essence of CNN lies in partitioning the feature space and allocating corresponding regions for known categories, without considering the presence of unknown categories; 3) Loss function design: The common cross-entropy loss functions neglect intra-class compactness in the feature space, resulting in cases where distances within the intra-class feature space are greater than inter-class distances.

This paper is dedicated to tackling the challenge of detector response when confronted with unseen category, namely open-set detection. Our aim is to train a detector using a closed set while endowing it with the ability to detect all known categories and identify unknown objects within an open set. In open-set detection, recognizing unknown object

This work was supported in part by the National Natural Science Foundation of China (Grant No. 62271153) and the Natural Science Foundation of Shanghai (Grant No. 22ZR1406700). (Corresponding author: Haipeng Wang)

Xiayang Xiao, Zhuoxuan Li,Haipeng Wang (e-mail: hpwang@fudan.edu.cn) are with the Key Laboratory for Information Science of Electromagnetic Waves (Ministry of Education), School of Information Science and Technology, Fudan University, Shanghai 200433, China.

categories presents the following challenges for conventional detectors: 1) The detector needs to provide accurate detection proposals simultaneously for known categories and potential unknown categories; 2) It is necessary to effectively leverage the model's learning of known categories to distinctly separate unknown categories from the background; 3) Modeling for objects of different sizes and simultaneous detection are required.

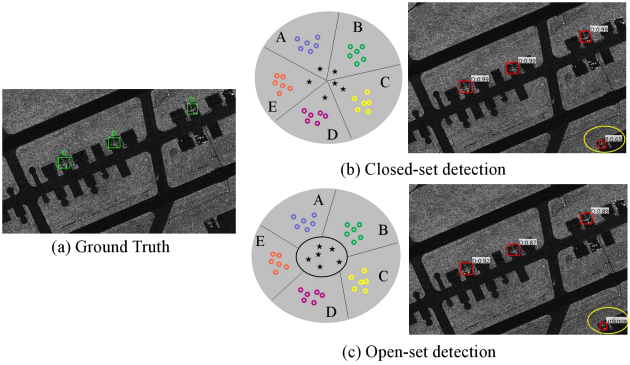

Fig.1 Compare closed-set and open-set detection results. (a) It shows the ground truth labeling for the test image, where only classes "A, B, C, D, E" are known in this experiment. (b) In closed-set object detection, such as Faster R-CNN, objects from unknown classes often tend to be overlooked or misclassified as known categories, as seen in category B within the yellow elliptical box in the lower-right corner. (c) In contrast to the objects within the yellow box in (b), our task objective is to localize and accurately recognize targets from unknown categories.

Recently, the field of computer vision has witnessed a surge in research aimed at addressing the challenge of detecting unknown objects. Among these endeavors, one of the most notable studies are conducted by Joseph et al.[11], introducing the towards open world object detection (ORE). This method leverages class-agnostic proposals generated by the region proposal network (RPN) to automatically acquire pseudo-unknown objects. Subsequently, it clusters these automatically labeled unknown objects with known categories, ultimately distinguishing between unknown and known categories through an energy distribution approach. While the research marks a significant milestone in open-set detection research, it still presents some problems: 1) The classical design of Faster R-CNN falls short in encoding long sequential dependencies. However, capturing contextual information in images is paramount for perceiving unknown objects. 2) Conventional RPN fundamentally operate as binary classifiers, and under strong supervision based on known categories, the generated proposals tend to favor known categories. This may lead to overfitting issues with the training data. In addtion, Han et al. [12] also propose an unknown object detector. Their approach employs a single potential region to represent unknown objects, which may result in inaccurate classification and suboptimal separation between known and unknown categories.

In order to better model unknown objects in SAR images and capture their spatial semantics, inspired by previous works [11]-[14], this work proposes OSAD, which is designed to facilitate the detection and recognition of unknown targets in SAR images under open environmental conditions. It is noteworthy that, while preserving the closed-set performance, the proposed approach has led to improvements in all open-set metrics. Specifically, an absolute gain of 18.36% in the average precision ($AP_u$) for unknown targets are acchieved in the OS-SAR-Aircraft dataset. This research holds significant implications for the application of intelligent algorithms in dynamically changing real-world environments.

In particular, this approach is equipped with dedicated components addressing the challenges posed by the open-set environment in SAR image analysis. These components encompass Global Context Modeling (GCM), Quality-Driven Pseudo Label Generation (LPG), and Prototype Contrastive Learning (PCL), aimed at effectively detecting unknown entities within SAR images. Inspired by the human recognition mechanism, this framework mimics human robustness when encountering unfamiliar data, often relying on templates or prototypes for unfamiliar entities. The GCM and PCL components aim to enhance the performance of the CNN extractor, analogous to simulating human sensory organs, transforming specific objects into abstract feature representations. The entire process of feature extraction emulates the human perceptual process. On the other hand, the PCL module learns prototypes for each known class, similar to abstract memories in the human brain corresponding to respective categories. Analogous to human cognition, the classification decisions within the PCL module are achieved by matching abstract features with prototypes of each known category. If the CNN features of a test sample do not align well with the prototypes of known categories, it is considered as unknown, mirroring the human cognitive process.

Our primary contributions can be summarized as follows:
(1) To the best of our knowledge, our study stands as the pioneering endeavor to apply an open-set setting to aircraft detection in SAR images, achieving an absolute gain in average precision for unknown targets from 0 to 18.36, allowing it to better adapt to real-world scenarios.
(2) This paper introduces a novel open-set SAR detector, featuring meticulously designed learners LPG and PCL, trainable end-to-end, and directly applicable in open-set environments.
(3) To enhance the detector's capability in identifying unknown entities, this study have improved traditional RPN networks, introducing a proposal network based on quality positioning clues, achieving generalization beyond categories and datasets, effectively enhancing perception of unknown entities.

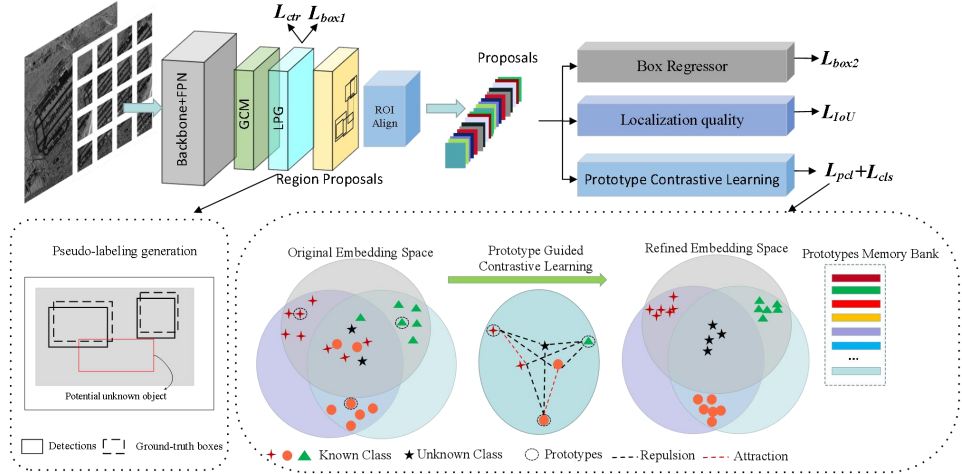

Fig. 2 It provides an overview of the OSAD architecture.

## II. RELATED WORK

### A. Open-set Recognition

In closed-set learning, it is typically assumed that only previously known classes exist during testing, while open-set learning assumes the possibility of known and unknown classes during the testing phase. Scheirer et al. [15] are the first to treat open-set recognition (OSR) as a minimization-constrained task. They develop open-set classifiers based on Support Vector Machines (SVMs), which is capable of rejecting unknown objects during testing. Subsequently, researchers extended the open-set framework using SVM-based methods for multi-class classifiers, as well as probabilistic models and extreme value theory classifiers to address the problem of disappearing confidence in unknown classes[16]-[18]. Bendale et al. achieve the first deep learning-based open-set recognition method, known as OpenMax. They model distances on activation vectors using Weibull distribution and re-calibrated the Softmax layer's probabilities to recognize unknown objects in deep network feature space[19]. Additionally, Ge et al. [20] propose a method building upon OpenMax that utilizes generative models to synthesize unknown samples for effective separation of known and unknown samples. OpenGAN [21] employs generative adversarial training to discriminate unknown objects, where generated latent open images assist in identification. Furthermore, some methods employ encoders [22][23] or reconstruction-based approaches [24], using reconstruction errors as indicators for unknown recognition. On the other hand, distance-based prototype discrimination methods [25] measure the distance between image features and learned prototypes for open-set image recognition. Our approach shares some similarities with [25], but differs in that this work encode contrastive features in prototype-based discrimination and employ momentum sequence updates for known class prototypes.

### B. Open-set object Detection

In the field of object detection, open-set detection can be seen as an extension of open-set recognition. It is worth noting that, despite extensive research in the domain of optical images, such research has not yet been conducted in the SAR image domain. Dhamija [26] is the first to analyze the impact of an open-set setting on classical detectors in his research. He finds that most detectors exhibited excessive confidence when misidentifying unknown class targets as known class targets. To enhance detection performance under open conditions, researchers like Miller [27][28] utilizes dropout techniques based on Bayesian inference [29] approximations of neural network parameters, aiming to assess label uncertainty. Additionally, Joseph [11] and his colleagues observe that under open-set conditions, the Helmholtz free energy for unknown classes is higher. Therefore, they propose to utilize energy measurements to determine whether a sample belongs to an unknown class. OW-DETR[30] introduces attention feature maps for scoring candidate proposals objectively and selecting the top K candidate boxes to generate pseudo-labels, which are then distinguished between known and unknown categories using a novelty classifier. However, it is important to note that the methods proposed by [11] require labeling of unknown classes, which contradicts the open-set detection setup. Past methods mainly relied on feature outputs (logits) from pre-trained models in the latent space to represent unknown classes. However, as pointed out by UC-OWOD[30], unknown objects may span multiple categories, making this representation method less accurate. To better distinguish between known and unknown objects, this paper attempts to use prototype contrastive learning to achieve compactness within the latent space and separation between different categories through distance discrimination.

### C. Prototype Models

The initial prototype model is known as Learning Vector Quantization (LVQ) [31], originating from the concept of k-Nearest Neighbors (KNN). LVQ preserves one or multiple prototypes for each category, which are refined through learning from the data to better represent it, thus effectively distinguishing between different categories. Additionally, various methods for prototype learning have been proposed, with some focusing on improving prototype update rules during training, such as [32]-[35]. Meanwhile, other approaches model the problem as a challenging parameter optimization task, guided by more intricate loss functions, to learn how to present prototypes[36]-[39]. However, these methods are often based on manually designed features. Subsequently, prototype learning is introduced into the field of CNN. For prototype networks, it is assumed that there exists an embedding space where data points gather around independent prototypes for each category. Snell et al.[40] propose incorporating the concept of prototypes into CNN for few-shot learning, achieving satisfactory results. Considering that unknown class objects typically involve few-shot scenarios, in this paper, this work suggests jointly learning prototypes and contrastive learning from the data. This work constructs a standard end-to-end deep framework to explore prototype distance-based methods for classifying unknown objects.

## III. MATHDOLOGY

### A. Problem Definition

In accordance with prior research[11][12], the concept of open-set detection within the OSAD task is defined as follows: The dataset $D$ is divided into the training data $D_{tr}$ and the testing data $D_{te}$. $D_{tr} = \{(x, y), x \in X_{tr}, y \in Y_{tr}\}$, $D_{te} = \{(x', y'), x' \in X_{te}, y' \in Y_{te}\}$, where $x$ and $x'$ signify the input images, while $y$ and $y'$ denote their respective annotations including class labels and bounding box coordinates. The training data $D_{tr}$ contains K known classes, $C = \{c_1 ... c_k\}$, whereas the testing data $D_{te}$ comprises both known and unknown instance $c_u \notin C$. An open-set detector is trained on $D_{tr}$ aiming to accurately detect all known class objects within $D_{te}$, while also effectively identifying unknown class objects to prevent misclassification as known classes.

### B. Overall Architecture

As depicted in Fig. 2, the proposed OSAD is a two-stage detector with three key modules: GCM, LPG, and PCL, designed to enhance open-set detection. GCM (Section III-C) enriches contextual information and improves feature encoding. It's helpful to mitigate inductive bias and potentially enhance detection accuracy by capturing long-range dependencies with multiscale receptive fields. LPG (Section III-D) addresses the challenge of identifying unknown classes within the background by using a quality-guided proposal network to filter candidate regions, as manually annotating infinite unknown classes is impractical. PCL (Section III-E) aids the detector in distinguishing between categories within the latent space through prototype contrastive learning, achieving clear differentiation between known and unknown objects by encoding the distance between proposals and known class prototypes. In the ensuing sections, we will delve into a meticulous explication of each of these integral components.

### C. Global Context Modelling

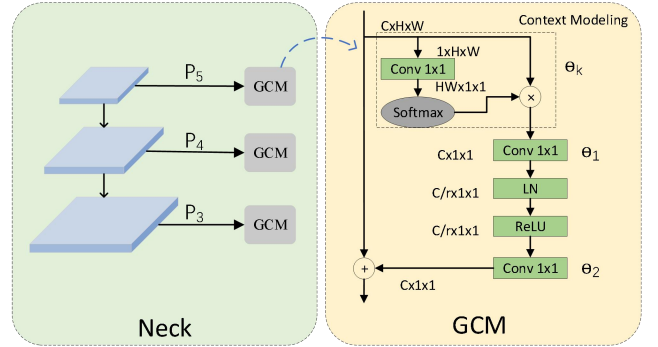

Fig. 3 Architecture of GCM. C, H, W denote the number of channels, height, and width of the feature map. r signifies the bottleneck ratio. ⊗ represents matrix multiplication. ⊕ indicates the broadcast element wise addition.

The capacity of the feature representation network to extract discriminative features is of paramount importance, particularly in downstream tasks involving robust target identification[41]. When considering the necessity for open-set detection, the scenario can grow in complexity due to the potential presence of diverse unknown objects within the images. Hence, it becomes imperative to ensure that the feature representation network possesses enhanced capabilities in encoding global contextual information effectively. This enhancement is essential to enable the network to accurately perceive objects that have not been encountered during training. In particular, when dealing with the prediction of objects at multiple scales, feature maps at distinct hierarchical levels exhibit remarkable sensitivity in capturing various fine-grained details and scale-related information of targets. Consequently, during the encoding of global contextual information, the incorporation of large receptive field information across multiple scales in the images becomes indispensable. This facilitates superior adaptability to targets of varying sizes. Furthermore, this refined detection framework serves to mitigate inductive biases and minimizes presumptions regarding unknown objects during the testing phase. As a result, it significantly contributes to the enhancement of detection performance.

To construct a global contextual framework, this work introduces the GCM [42] for SAR aircraft detection. The incorporation of the GCM empowers us to acquire contextual information from the images and establish long-range dependencies, resulting in the creation of a more robust multi-scale feature representation. As illustrated in Fig. 3, this work integrates the feature maps Pi (i=3,4,5) obtained from the neck with the GCM module. The GCM module comprises the following key components: (a) Global attention pooling: This component obtains attention weights through $1 \times 1$ convolution and a Softmax function. Subsequently, these weights are employed for attention aggregation, enabling the extraction of global contextual features. This step helps mitigate potential information loss that may occur due to dimensionality reduction. (b) Bottleneck transformation: Designed to capture dependencies between channels, this aspect utilizes $1 \times 1$ convolution for feature transformation. (c) Element-wise addition: This process is utilized to seamlessly integrate global contextual features into the features at each position. Furthermore, this work introduces layer normalization to alleviate optimization challenges that arise as a result of the bottleneck transformation. Through this methodology, the proposed model can effectively capture global information present in SAR images, thereby enhancing the performance of aircraft detection. The entire process is thoughtfully modeled to facilitate these improvements.

$$Z_i = \theta_2 \text{ReLU}\left(\text{LN}\left(\theta_1 \sum_{j=1}^{N_p} \alpha_j\right)\right) + X_i \quad (2)$$

$$\alpha_j = \frac{e^{\theta_k X_j}}{\sum_{i=1}^{N_p} e^{\theta_k X_n}} \quad (3)$$

X and Z symbolize the input and output feature vectors, respectively. The $X_i$ denotes information pertaining to the presently attended position, while $X_j$ signifies global information, with j iterating over all conceivable positions. $N_p$ is indicative of the number of positions within the feature map, which, in the case of an image, stands as $N_p = H \times W$. The symbols $\theta_k$, $\theta_1$ and $\theta_2$ represent linear transformation matrices, conventionally achieved via $1 \times 1$ convolutional operations. $\alpha_j$ stands as the weight associated with global attention pooling.

*D. Localization Quality-Driven Pseudo label Generation*

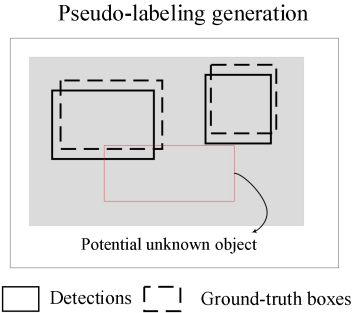

Fig.4 The purpose of LPG is to make the RPN network better perceive unknown targets and generate higher quality unknown target proposal boxes (Red box is a potential unknown target)

In order for the detector to be able to detect unlabeled unknown objects in the training set, the open detection framework needs to rely on the presence of potentially unknown instances in the training images, which involves labeling these unknown objects using true unknown classes. However, it is impractical to re-annotate all instances of every image in an annotated dataset. ORE utilizes an automatic labeling step to obtain pseudo-unknown objects for training. Automatically labeled pseudo-labels refer to the proposal boxes associated with the RPN output with potentially unknown objects.

However, standard RPN is trained under strong supervision of known categories, and the proposal boxes it generates may be biased towards known categories. Inspired by [14], without category label supervision, this study proposes LPG, which learns to use cues from object location and shape (called localization quality) to enhance generalization to unknown object proposals. LPG is designed to enable RPN networks to generalize better to new and unseen categories. As shown in Fig. 4, the pseudo-labels in LPG refer to proposal boxes with high objective scores that do not overlap with ground truth (GT) known instances.

Specifically, in standard RPN, the input is the output features of each level of the feature pyramid. Each feature map is processed through a separate convolutional layer, and finally two independent sub-layers are formed, one for performing bounding box regression and the other for classification tasks. Different from the standard RPN design, in the first stage, this work chooses to replace the classification branch with the centrality regression branch [44]. Because in the anchor frame proposal stage, the learning of target positioning is more critical than classification. This design helps avoid the problem of overfitting to foreground categories, thereby promoting generalization to unknown categories. For box regression, this work applies the distance ( left, right, top, bottom, lrtb ) from the position to the four sides of the real box to calculate the regression loss. In the second stage, the high-scoring proposal boxes filtered out in the RPN are used to perform RoIAlign, and then enter the refinement regression of the bounding box and the IoU regression head [45]. When evaluating the quality of the target proposal box, this work adopts the dual measurement of centrality score and IoU score. The IoU score not only helps the model refine the proposal score, but also helps avoid the model's overfitting of the foreground category. The objective score of a proposal region is calculated as the geometric mean of centrality $c$ and IoU score $b$, $s = \sqrt{c \cdot b}$. Finally, the proposed boxes with high objectivity scores that do not overlap with ground-truth objects are marked as potential unknown instances.

During the training process, if the IoU between a proposal and the corresponding ground truth box is greater than the threshold 0.7, it is judged as a positive sample; if the IoU is less than 0.3, it is judged as a negative sample. Subsequently, 256 selected anchor frames are randomly sampled for loss calculation. The loss formula of LPG can be expressed as follows:

$$L_{LPG} = \lambda_1 L_{box1} + \lambda_2 L_{ctr} + \lambda_3 L_{box2} + \lambda_4 L_{IoU} \quad (4)$$

Here, $L_{box1}$, $L_{ctr}$, $L_{IoU}$ and $L_{box2}$ represent lrtb bounding box regression, centrality regression, $xywh$ bounding box regression ($(x,y)$, weight, height, $xywh$) and IoU regression, respectively. Meanwhile, $\lambda_1$, $\lambda_2$, $\lambda_3$ and $\lambda_4$ denote the corresponding weight coefficients. The first two loss functions are utilized during the initial proposal generation, whereas the latter two loss functions play a role in the proposal refinement process.

*E. Prototype Contrastive Learning*

The classic Faster R-CNN [3] approach involves the integration of proposals into fixed-sized layers, subsequently encoding them as RoI features. Nevertheless, in certain scenarios, training reliable representations for unknown classes using known samples becomes a formidable challenge. Consequently, there is a need for a classification methodology capable of distinguishing features among different categories to elevate classification accuracy. This section delves into strategies for enhancing recognition performance by mitigating the overlap between features extracted from known-class samples and those from unknown samples.

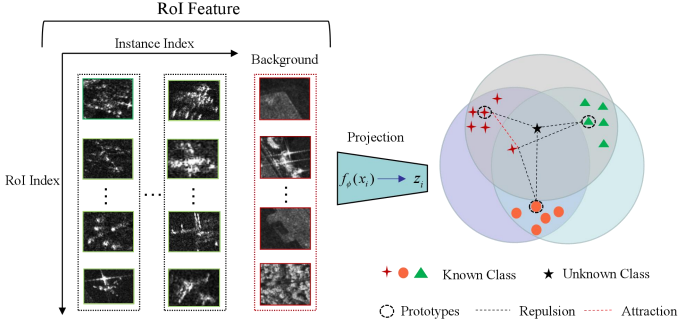

Fig.5 The illustration of prototype contrastive learning. The projection head is used to encode RoI features into the embedding space to compute the PCL loss, the parameters of which are updated during the training phase. The prototype contrastive loss promotes instance-level intra-class compactness and inter-class variance, mitigating the issue of misclassification.

Considering that only few labels are used in semi-supervised learning to learn better and more accurate object region representations, this study focuses on contrastive learning[46]. This learning paradigm empowers the model to map positive samples to closely aligned representations while concurrently driving negative samples apart within the embedding space. Such an approach holds the promise of augmenting the model's prowess in representation learning. Indeed, contrastive learning methods have exhibited conspicuous advantages in contemporary downstream classification tasks.

As illustrated in Fig.5, our approach begins by mapping the input data into an embedding space through a non-linear transformation [41]. Within this embedding space, class prototypes are treated as the centroids of training sets. Then, classification is performed on embedded query points by finding the closest prototype. Therefore, this work leverages contrastive learning techniques to enhance category separation within the latent space. Contrastive learning encourages instances of the same class to remain closely clustered while pushing instances from different classes apart. This separation of known classes significantly contributes to the recognition of unknown classes.

Specifically, this work utilizes 1024-dimensional feature vectors obtained from the RoI head. These feature vectors $f_\Phi(x_i)$ are projected into a lower-dimensional embedding vector $z_i$ via a projection head. The projection head is composed of a non-linear fully connected layer connected to the RoI. It includes fully connected (FC) layers, ReLU activation layers, and additional FC layers. It's important to note that the encoder, which includes this projection head, is used exclusively for the classification branch during training and is not employed during inference. Throughout the training process, the prototype contrastive loss is applied to the classification branch. During downstream tasks, this work adopts the feature vectors from $f_\Phi(x_i)$ instead of those obtained by the projection head.

The prototypes which calculate by encoder $f_\Phi$, are represented by M-dimensional vectors and denoted as $p_k \in R^M$, and possess learnable parameters $\Phi$. Each prototype serves as the average vector of embedded support points belonging to its respective category. The prototypes for each class can be likened to abstract memories of that class in the human brain.

$$p_k = \frac{1}{|C_k|} \sum_{(x_i,y_i) \in C_k} f_\Phi(x_i) \tag{5}$$

Prototype contrastive loss measures the similarity between pairs of samples in a representation space. This study defines the prototype contrastive loss as follows:

$$L_{pcl} = \sum_{k=0}^{K} l(z_j, p_k) \tag{6}$$

$$l(z_j, p_k) = \begin{cases} \Delta(z_j, p_k), j = k \\ max\{0, T - \Delta(z_j, p_k)\}, j \neq k \end{cases} \tag{7}$$

In the formula (7), $\Delta$ represents the cosine distance between the embedding $z_j$ and the prototype $p_k \in P$, while T is a hyperparameter used to denote the minimum distance between input feature vectors and prototypes of different classes in the latent space. By minimizing this loss, it ensures a good separation between different classes in the latent space. In addition, the performance of closed sets in the potential space is not as good as the traditional softmax classifier, so the softmax classifier is still used for known objects. The cross entropy loss $L_{cls}$ is employed to optimize the embedding vector, the same as Faster-rcnn.

| Algorithm 1: Prototype Contrastive Loss | |
|---|---|
| Input： | Input Embedding Vector $z_j$ <br> Class Prototype Vector Library: $P = p_1, p_2, \ldots, p_k$ <br> Hyper-parameter: Iterations $I_m = 500, I_n = 1000$，<br> Momentum factor $\eta$ |
| Output： | Contrastive Loss: $L_{pcl}$ |
| Main loop: | |
| 1: | Initialize $P$ if it is the first iteration |
| 2: | $L_{pcl} \leftarrow 0$ |
| 3: | **If** iteration i == $I_m$ then |
| 4: |    Update $P$, Eigenvector averages for each class in $F_{stroe}$ |
| 5: |    Update $L_{pcl}$, Calculate the $L_{pcl}$ by equation.6 |
| 6: | **else if** iteration i > $I_m$ then |
| 7: |    **if** i % $I_n$ == 0 then |
| 8: |      $P_{new} \leftarrow 0$ Eigenvector averages for each class in $F_{stroe}$ |
| 9: |    Update $P$, $P = \eta P + (1 - \eta)P$ |
| 10: |    Update $L_{pcl}$, Calculate the $L_{pcl}$ by equation.6 |
| 11: | **return** $L_{pcl}$ |

Class prototypes are dynamically updated through the iterative averaging of feature vectors stored in corresponding queues. To capture the evolving nature of feature vectors during training, this work stores class-specific features within their respective queues, denoted as $F_{stroe} = \{q_0, q_{1,\ldots,} q_C\}$, where $q_k$ represents the queue length for each class. This is a scalable method for tracking how feature vectors evolve with training. In addition, the number of feature vector stored is limited by $C \times Q$, where $Q$ signifies the maximum size of the queue. Algorithm 1 provides insight into the procedure for updating class prototypes while calculating the prototype contrastive loss. The prototype contrastive loss is integrated into the standard detection loss and is backpropagated to train the network end-to-end.

## IV. EXPERIMENTAL RESULTS AND ANALYSIS

### A. Datasets

This work has created a dataset specifically tailored for open-set aircraft detection in SAR imagery, denoted as OS-SAR-Aircraft. This dataset encompasses a total of 57 large-scale SAR images sourced from China's GF-3 satellite. The satellite operates within the C-band and utilizes both HH and VV polarization modes, employing the SpotLight (SL) observation mode to capture large-scale images with a nominal resolution of 1.0 meter. The meticulous annotation of aircraft within these SAR images are carried out by experts. Their annotation process involves a fusion of domain expertise and the integration of relevant optical imagery to ensure precision. The dimensions of these images span from 12,000×8,000 pixels to 20,000×20,000 pixels.

The entire dataset encompasses eight distinct categories, designated as A, B, C, D, E, F, G, and H. This work adopts a random selection of 40 images to constitute the training set, denoted as $D_{tr}$, while the remaining images form the test set, referred to as $D_{te}$. To facilitate training procedures, particularly for large-sized images, this work segments them into $1024 \times 1024$-pixel patches, with an overlap of 200 pixels between adjacent patches. This preprocessing step ensures the dataset's suitability for training models.

In addition, this study has devised seven distinct tasks using this dataset. In Task T-1, both the training and test sets exclusively comprise known instances, encompassing five categories (A, B, C, D, E). The purpose of this task is to evaluate the closed-set model, meaning it is tested without any unknown objects. In Tasks T-{2,3,4}, similar to Task T-1, there are no labels for unknown instances in the training set. However, the test set gradually introduces open-set categories, including five known classes and {1,2,3} unknown classes. In Tasks T-{5,6,7}, the training set configuration mirrors that of Task T-1, but the test set progressively increases the Wilder Ratio (WR)[28], comprising 40 test images with only known categories and {44,70,80} images with unknown categories that do not intersect with the categories in the training set. For detailed information, please refer to TABLE I.

### B. Evaluation Metrics

This work utilizes the Wilder Impact (WI) [26]and Absolute Open Set Error (AOSE) [28] metrics to assess the performance in open-set scenarios. The Wilder Impact metric is employed to evaluate the degree to which unknown objects are misclassified as known classes. On the other hand, AOSE calculates the number of instances where unknown categories are erroneously categorized as known categories. These two metrics implicitly evaluate the model's effectiveness in dealing with unknown objects.

$$\text{Wilderness Impact (WI)} = \left(\frac{P_k}{P_{k \cup u}} - 1\right) * 100\% \tag{7}$$

$P_k$ and $P_{k \cup u}$ represent the accuracy of known categories under the closed-set and open-set conditions, respectively. For the sake of simplicity, this work has scaled up the original Wilder Impact by a factor of 100. This study reports the Wilder Impact at a confidence level of 0.5. While assessing the open-set detection capabilities of the detector, this study also monitors changes in its closed-set detection performance. Therefore, this study simultaneously reports the $mAP_k$ of known categories to evaluate closed-set object detection performance. Additionally, for a more intuitive quantification of the ability to detect unknown objects, this study also reports the $AP_u$ of unknown categories. It's important to note that WI, $AOSE$, and $AP_u$ are open-set metrics, whereas $mAP_k$ is a closed-set metric.

TABLE I The components of the OSAD task, as well as the quantity of images and object instances within each task.

| Split | T-1 | T-2 | T-3 | T-4 | T-5 | T-6 | T-7 |
|---|---|---|---|---|---|---|---|
| Class | 5 | 5-6 | 5-7 | 5-8 | 5-6-Mix | 5-7-Mix | 5-8-Mix |
| Training Images | 160 | 160 | 160 | 160 | 160 | 160 | 160 |
| Testing Images | 40 | 89 | 100 | 130 | 84 | 110 | 120 |
| Training Instances | 830 | 830 | 830 | 830 | 830 | 830 | 830 |
| Test Instances | 170 | 598 | 681 | 826 | 401 | 499 | 510 |

### C. Implementation

This experiments are run on PyTorch 1.9.1 and CUDA 11.1. The Faster R-CNN model is constructed using the Detectron2 (version 0.5) framework[48]. This work selects the ResNet50 as backbone [49] and integrate it with a feature pyramid network. For optimization, this work employs the SGD optimizer with an initial learning rate of 0.02, a momentum of 0.9, weight decay of 0.0001, and a maximum of 20,000 iterations. To adjust the learning rate, this work implements a cosine annealing strategy at 7,000 and 12,000 iterations.

All model training processes are completed on four NVIDIA GeForce RTX 3090 GPUs, with a batch size of 16. For the OS-SAR-Aircraft dataset, this work configures a total of 9 classes, including the 9th class designated for the background category. In the GCM phase, the chosen bottleneck ratio ($r = 4$) is primarily applied to the feature pyramid's P3, P4 and P5 stages.

In the LPG phase, this work sets the number of proposal boxes ($Pro_{num}$=256), the threshold used to define positive and negative sample boxes ($Pos_{iou}$=0.7), and he proportion used to calculate the positive sample loss ($Pos_{ratio}$=0.5) to optimize the center regression and bounding box regression losses.

Additionally, this work configures $Pro_{num}$ as 512, $Pos_{iou}$ as 0.5, and $Pos_{ratio}$ as 0.25 to optimize the IoU regression loss and xywh regression loss. Based on experimental insights, this work assigns values of $\lambda_1, \lambda_2, \lambda_3$ and $\lambda_4$ as 1, 8, 1, and 2, respectively. Concerning the PCL component, the queue size is set to 16, and T is set to 13.

During the training process, the LPG phase generates 2000 proposal boxes for scoring, which are utilized in subsequent proposal handling. During the inference stage, this work selects the top 1000 proposal boxes based on their scores and subsequently apply non-maximum suppression (NMS) with an IoU threshold set to 0.7. Following this step, this work performs RoIAlign and proposal box refinement. After filtering out proposals with confidence scores less than 0.05, an object score (s) is computed for each remaining proposal. These proposal boxes are then encoded into the latent space to calculate distances between each prototype. If the minimum distance between all prototypes exceeds a predefined threshold T, the proposal is classified as an unknown object. In Section 4.6, this study provides a comprehensive examination of the hyperparameters.

*D. Experiment of Comparison*

To comprehensively assess the performance of our proposed method, OSAD is being compared against several state-of-the-art techniques, including Faster RCNN (FR-CNN)[3], Dropout Sampling (DS)[27], OW-DETR[30], PROSER[48], and OpenDet[12]. We exclude ORE from the comparison since it relies on a validation set with annotations for unknown objects, which goes against the initial definition of open-set detection. FR-CNN is chosen to evaluate changes in performance for known categories. DS, a pioneering method, extends dropout sampling techniques to object detection. It estimates uncertainty information for labels by assessing Bayesian object detection systems on datasets, thereby enhancing detection performance under open-set conditions. PROSER leverages an additional class placeholder to extend the closed-set classifier, acting as a threshold boundary between known and unknown categories. It calibrates overly confident predictions for unknown categories by retaining the placeholder. OpenDet identifies unknown objects based on the assumption that they typically occupy low-density regions. It separates high-density and low-density regions in the latent space to identify unknown objects. To ensure a fair and consistent comparison, this work adopts the settings for all compared methods as referenced from their respective original papers while maintaining identical combinations of training and testing samples. According to the settings in Section IV-C and TABLE I, the experimental results are presented in TABLE II.

TABLE II. presents a detailed performance comparison, assessing the impact on various tasks by gradually introducing unknown categories. To begin with, in Task T-1, this work conducts a closed-set performance validation, which serves as a benchmark test without performance metrics related to open-set scenarios. It is worth noting that Faster R-CNN failed to detect unknown targets in all tasks, resulting in an $AP_u$ consistently at 0, highlighting the challenging nature of open-set detection. However, the proposed OSAD, by enhancing the detection and classification performance of unknown objects, achieved the best $AP_u$ performance in Tasks T-2 to T-4. This underscores OSAD's effectiveness in discovering and identifying unknown objects. Furthermore, it is noteworthy that, with the capability for detecting unknown objects in place, the OSAD also demonstrated an improvement in $mAP_k$. In summary, the results emphasize the effectiveness and superiority of the OSAD in open-set detection tasks, as well as its positive impact on the performance of known classes. These findings provide strong support and reference for research and applications in the field of open-set detection. The following parameters are reported at a confidence level of 0.5.

TABLE II Comparison with Other Methods. The performance of closed-set detection ($mAP_k$) is shown in T-1, and the performance of different methods on closed-set ($mAP_k$) and open-set (WI, AOSE, $mAP_k$, $AP_u$) is reported in T-{2,3,4} tasks. Open-set performance metrics quantify the model's ability to retrieve unknown objects.

| Methods | T-1 | T-2 | | | | T-3 | | | | T-4 | | | |
|---|---|---|---|---|---|---|---|---|---|---|---|---|---|
| | $mAP_k\uparrow$ | WI$\downarrow$ | AOSE$\downarrow$ | $mAP_k\uparrow$ | $AP_u\uparrow$ | WI$\downarrow$ | AOSE$\downarrow$ | $mAP_k\uparrow$ | $AP_u\uparrow$ | WI$\downarrow$ | AOSE$\downarrow$ | $mAP_k\uparrow$ | $AP_u\uparrow$ |
| FR-CNN | 94.24 | 0 | 35.00 | 96.67 | 0 | 4.48 | 55 | 95.47 | 0 | **2.68** | 96.00 | 95.76 | 0 |
| PROSER | 94.25 | 2.79 | 43.00 | 95.85 | 5.41 | 5.75 | 64.00 | 94.42 | 8.13 | 7.20 | 105.00 | 93.61 | 9.52 |
| Ow-DETR | 94.37 | 3.22 | 38.00 | 96.23 | 2.54 | 4.65 | 40.00 | 95.21 | 3.63 | 4.93 | 83.00 | 95.35 | 5.84 |
| DS | 94.52 | **1.41** | 41.00 | 96.04 | 1.50 | 6.55 | 65.00 | 94.96 | 2.37 | 5.60 | 105.00 | 94.85 | 2.91 |
| Open-Det | 91.93 | 4.50 | **27.00** | 93.36 | 1.22 | 8.51 | 38.00 | 90.51 | 2.33 | 7.92 | 62.00 | 91.17 | 2.66 |
| **OSAD** | **94.83** | 1.95 | 32.00 | **96.86** | **6.53** | **4.40** | **32.00** | **96.39** | **10.30** | 4.61 | **59.00** | **96.11** | **11.95** |

TABLE III Comparison with Other Methods in Task T-{5,6,7}

| Methods | T-5 | | | | T-6 | | | | T-7 | | | |
|---|---|---|---|---|---|---|---|---|---|---|---|---|
| | WI$\downarrow$ | AOSE$\downarrow$ | $mAP_k\uparrow$ | $AP_u\uparrow$ | WI$\downarrow$ | AOSE$\downarrow$ | $mAP_k\uparrow$ | $AP_u\uparrow$ | WI$\downarrow$ | AOSE$\downarrow$ | $mAP_k\uparrow$ | $AP_u\uparrow$ |
| FR-CNN | 0 | 40.00 | 96.31 | 0 | 8.00 | 86.00 | 94.32 | 0 | **6.80** | 108.00 | 94.61 | 0 |
| Proposer | 2.80 | 49.00 | 94.43 | 8.35 | 10.32 | 94.00 | 92.98 | 14.28 | 11.39 | 110.00 | 92.40 | 17.27 |
| OW-DETR | 4.13 | 44.00 | 95.78 | 5.73 | 9.78 | 88.00 | 94.37 | 7.83 | 8.64 | 93.00 | 94.32 | 12.43 |
| DS | 2.13 | 43.00 | 95.41 | 3.69 | 11.54 | 89.00 | 90.59 | 8.37 | 13.66 | 116.00 | 86.79 | 8.38 |
| Open-det | 6.04 | 35.00 | 86.05 | 3.07 | 14.63 | 62.00 | 84.60 | 6.65 | 13.58 | 71.00 | 84.68 | 7.36 |
| **OSAD** | **1.45** | **21.00** | **96.53** | **9.15** | **7.86** | **57.00** | **95.55** | **16.56** | 8.16 | **64.00** | **94.52** | **18.36** |

In order to provide a comprehensive evaluation of the detector's performance under various open-set conditions, this work has also introduced the WR. WR serves as an indicator of the proportion of unknown class objects relative to known class objects within the test set, offering insights into the challenges the model faces when encountering unknown factors. By incrementally increasing the value of WR, we can observe how the performance of the detector is influenced by unknown factors under various open-set conditions. As evident from the results in TABLE III, with the increase in WR, OSAD demonstrates a significant improvement in its ability to detect unknown targets. Particularly noteworthy are the achievements in tasks T-{5,6,7}, where OSAD attains $AP_u$ scores of {9.15, 16.56, 18.36}, signifying a substantial improvement compared to the performance outlined in TABLE II. This indicates that OSAD effectively distinguishes between known and unknown categories, adapting to the challenges posed by various open-set conditions. These results emphasize the robustness and versatility of the OSAD method, as it excels not only in closed-set tasks but also performs well under open-set conditions. This is of significant importance for real-world object detection tasks that need to address various unknown scenarios, such as in monitoring, security, and rescue applications. The successful application of OSAD will contribute to enhancing object detection performance across a range of applications, increasing sensitivity to unknown categories and providing a powerful solution for a broader spectrum of open-set problems.

Furthermore, upon a meticulous examination of the two tables presented above, this study can derive the following valuable insights:

1) Most open-set detection methods experience a general decline in their performance on closed-set tasks ($mAP_k$) as WR increases. The magnitude of this performance decrease to some extent reflects the adaptability of the detector to open-set scenarios. Typically, a smaller decrease in performance indicates greater stability in the model's ability to detect unknown categories.
2) An increase in WR implies that test images contain more objects from entirely unknown categories. By comparing the results of tasks T-{2,3,4} and T-{5,6,7}, it can be observed that the detector performs better in detecting $U_u$ (objects entirely unknown to the detector) compared to $U_k$ (objects that have been systematically trained but are treated as background $U_k$). This suggests that during the training process, the detector may have become aware of objects designated as background, hindering the detection of unknown targets $U_k$ This will be the focus of our next research steps.
3) A notable correlation exists between open-set and closed-set detection metrics. In general, detectors that excel in open-set performance often demonstrate commendable performance in closed-set conditions, but not vice versa. For example, open-set performance methods such as OSAD and PROSER exhibit relatively superior closed-set performance, often surpassing the majority of closed-set methods. However, DS and OW-DETR outperform PROSER in closed-set performance but cannot compete in open-set scenarios.
4) Common closed-set methods typically struggle to effectively adapt to open-set detection tasks. Open-set methods can perceive unknown categories while maintaining strong performance on known categories. Therefore, open-set detection methods hold greater practical value in real-world applications, as they can accommodate ideal closed-set assumptions while addressing the challenges of open-world detection. This provides more research and application opportunities in the field of open-set detection.

*E. Ablation Study*

This work conducts ablation experiments to analyze the impact of key components and core design choices on tasks T-1 and T-3. The overall analysis is as follows, with detailed comparisons provided in TABLE IV. We progressively integrate modules to quantify their influence on the detector's performance. In comparison to Scheme A, both Schemes B, C, and D demonstrate certain improvements. When compared to Scheme C, Scheme D exhibits a minor decrease in $mAP_k$, due to the influence of determining unknown categories on known category classification. However, compared to Schemes B and C, the overall performance enhancement observed in Scheme D underscores the advantages of module combinations. In summary, Scheme D outperforms Scheme A comprehensively, elevating $AP_u$ from 0 to 10.30. To provide a more intuitive representation of its effectiveness, specific detection performance comparisons are presented in Fig 9. From left to right are the ground truth, detection results of baseline, and detection results of OSAD.

TABLE IV The impact of gradually incorporating components into the baseline

| | GCM | LPG | PCL | T-1 | | | T-3 | |
|---|---|---|---|---|---|---|---|---|
| | | | | $mAP_k\uparrow$ | $WI\downarrow$ | $AOSE\downarrow$ | $mAP_k\uparrow$ | $AP_u\uparrow$ |
| A | | Baseline | | 94.24 | 4.48 | 55.00 | 95.47 | 0 |
| B | √ | | | 94.29 | **4.36** | 53.00 | 96.40 | 0 |
| C | √ | √ | | 94.51 | 5.72 | 41.00 | **96.52** | 0 |
| D | √ | √ | √ | **94.83** | 4.40 | **32.00** | 96.39 | 10.29 |

GCM: Learning global context information is crucial for capturing long-range dependencies between targets and backgrounds in SAR image scenes. In scenarios with various unknown objects present in the images, the integration of GCM allows the network to better perceive these potential targets. In tasks T-1 and T-3, we observe a significant improvement in performance. This indicates that the attention maps representing long sequential position relationships generated by the GCM effectively enhance the network's ability to represent objects of interest.

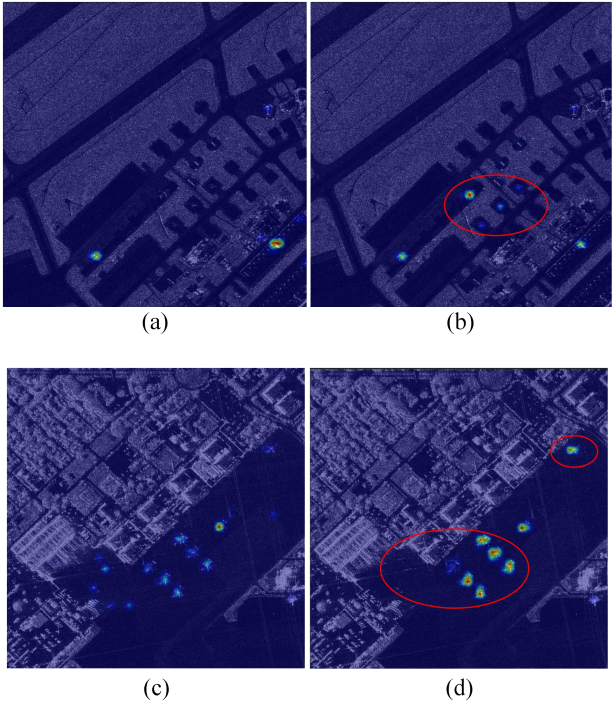

Fig. 6 In the qualitative analysis, this work presents partial heatmaps from the test set. (a) and (c) depict the detection heatmaps for the baseline, while (b) and (d) illustrate the detection heatmaps for OSAD. Please pay particular attention to the activated instances within the red elliptical boxes.

LPG: The design of the LPG network is aimed at enhancing the generalization capability of proposals for unknown objects. In contrast to the traditional RPN stage, LPG employs a different approach by eliminating the fixed pattern of image classification. It relies on the localization and shape cues of targets, specifically the quality of localization, to better capture the objects within the proposed regions. As illustrated in Fig.6, this work provides a comparison between baseline result images and heatmaps from Scheme C: (a) and (c) represent the baseline results, while (b) and (d) represent the results from Scheme C. In these examples, regions containing foreground targets exhibit higher heatmap activations. Compared to (a) and (c), it can be observed that in (b) and (d), more potential target regions are activated. In the first row of (b), some less conspicuous but aircraft-like regions have been activated. In the second row of (d), more aircraft targets are distinctly activated, whereas in (c), the brightness of activations in more potential target regions is lower. This suggests that LPG can assist the network in better perceiving regions containing targets, reducing potential biases that may arise during supervised training on known categories. In (d), a greater number of perceptual regions belong to unknown targets, and specific ground truth information can be found in Fig.7.below.

PCL: The incorporation of the PCL module has notably enhanced the performance of Scheme D compared to the baseline when dealing with unknown categories. It has elevated the performance from an initial score of 0 to 10.29, achieving this significant improvement while concurrently maintaining the performance for known categories. Additionally, during the transition from Scheme C to Scheme D, there have a slight decline in $mAP_k$ performance due to the introduction of the PCL module. This decrease can be attributed to the different distance discrimination method employed by the PCL module compared to traditional strong supervised classification. However, in the latent space, there exist objects that closely resemble known categories and unknown categories, which may lead to a minor decrease in the classification performance of known categories. Nonetheless, it is gratifying to note that Scheme D has exhibited a notable overall performance improvement compared to the baseline. To provide a clearer illustration of the discriminative feature space constructed by PCL, this work has employed the T-SNE dimension reduction visualization method for an intuitive presentation of its advantages on both known and unknown categories. The visualized results are displayed in Fig. 8.below.

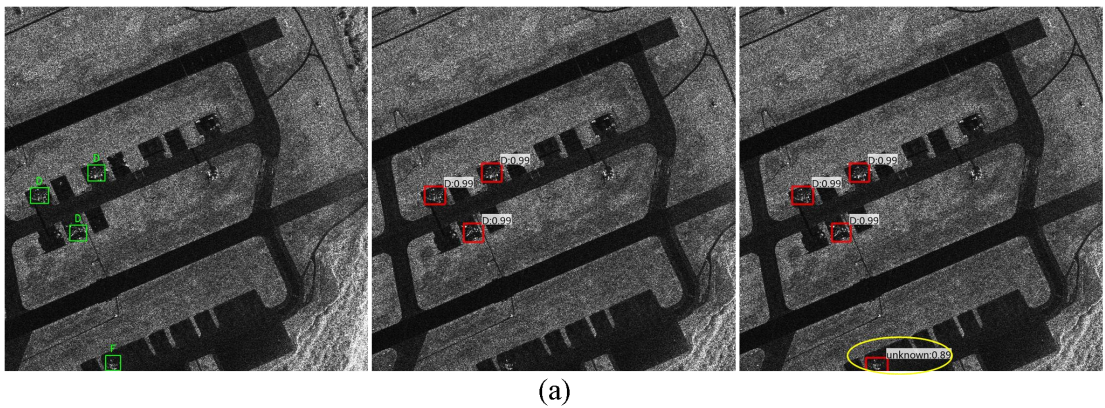

(a)

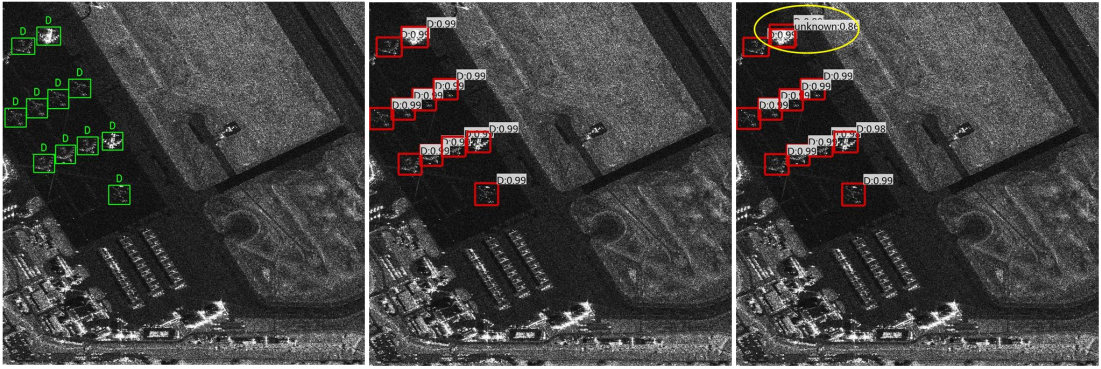
(b)

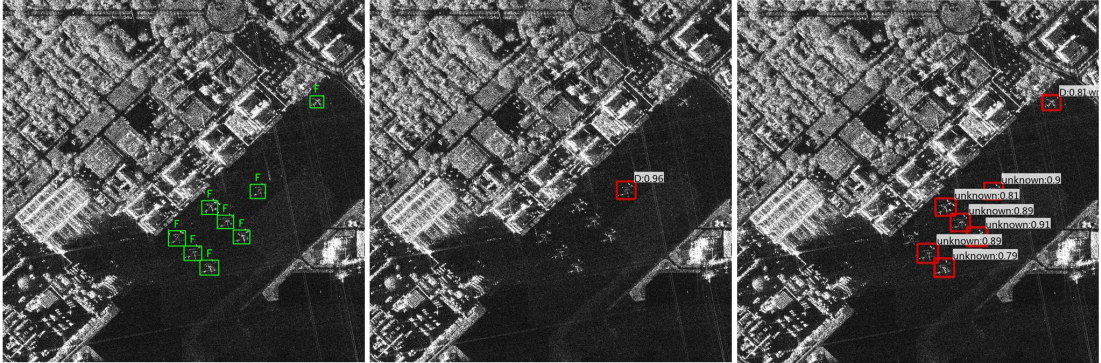
(c)

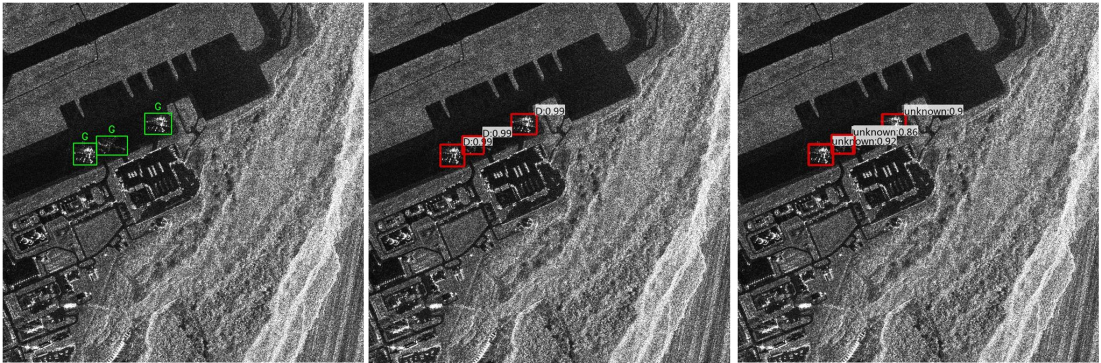
(d)

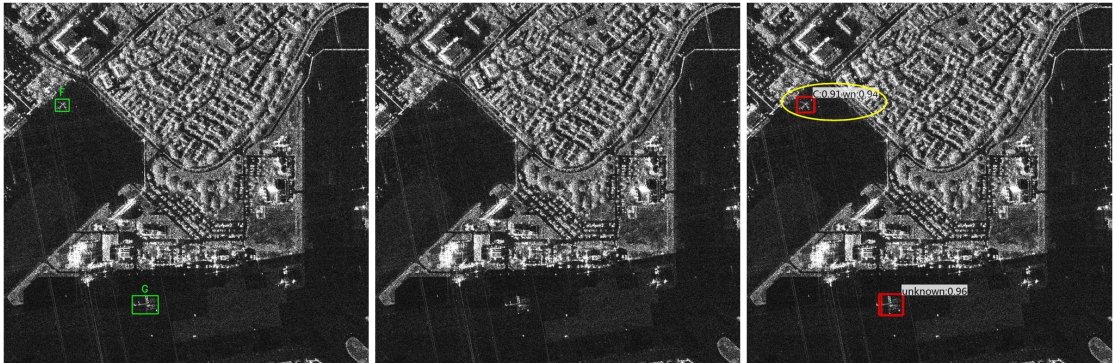
(e)

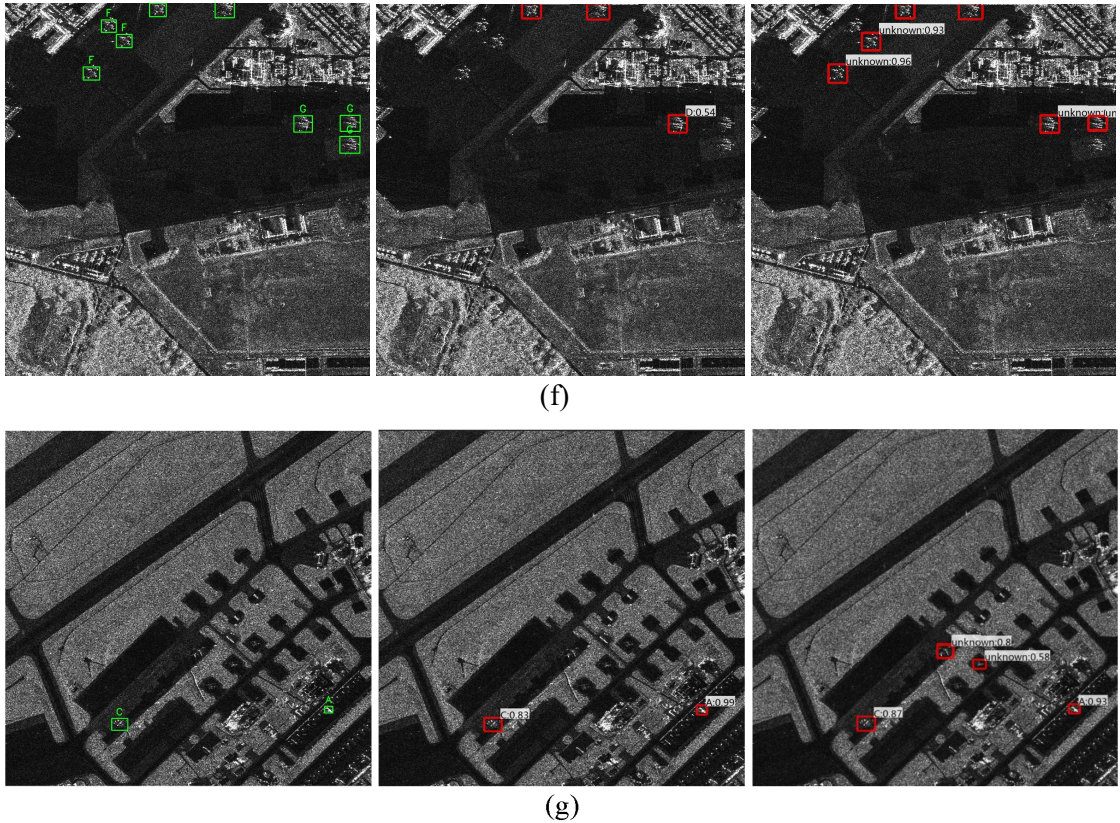

(f)

(g)

Fig.7 Ground truth (left). Qualitative comparison of the baseline (middle) and OSAD (right). In comparison to the baseline, OSAD indeed improves the detection of unknown objects. For example, in (a), unknown categories are omitted in the baseline, yet OSAD successfully detected unknown targets (highlighted within yellow elliptical boxes). In (b), all known targets are detected by both methods. The OSAD method within the yellow ellipsoid encompasses two boxes, with higher confidence assigned to known categories and lower confidence assigned to unknown categories, without impacting its closed-set performance. In (c), the majority of the unknown targets were detected by OSAD, while the baseline identified only one target, which is classified as known category D. In (d), all unknown objects are detected by OSAD, but they are all classified as known category D by the baseline. (e) and (f) demonstrate that the majority of targets are missed by the baseline, while OSAD consistently detected and identified unknown targets. In (e), one target is identified as known category C by OSAD, but the object is also assigned to the unknown category with a higher confidence. In (g), the baseline accurately detected known class targets, but OSAD detected potential unknown classes and correctly assigned corresponding labels.

In Fig.8(a), a notable deficiency in the separability of the feature space is observed, primarily characterized by the indistinct and overlapping boundaries between unknown and known categories. Additionally, the distribution of unknown classes exhibits a chaotic trend, thereby increasing the complexity of open-set image classification. However, through training with prototype contrastive learning, OSAD enhances the separability of the feature space by mapping different instances into a high-dimensional feature space. As exemplified in Fig.8(b), the feature distances between unknown and known categories undergo a significant amplification. This substantial refinement of the feature space bestows a multitude of advantages. Firstly, it elevates the discriminative power between known and unknown categories, thereby contributing to the enhancement of the classification performance of known categories. Secondly, by measuring the distance between the to-be-classified instance and known class prototypes, OSAD can effectively reject unknown categories, consequently reducing false positive rates in open-set environments. This unique feature space transformation strategy effectively addresses the challenges of open-set object detection, providing a solid theoretical foundation for performance enhancement.

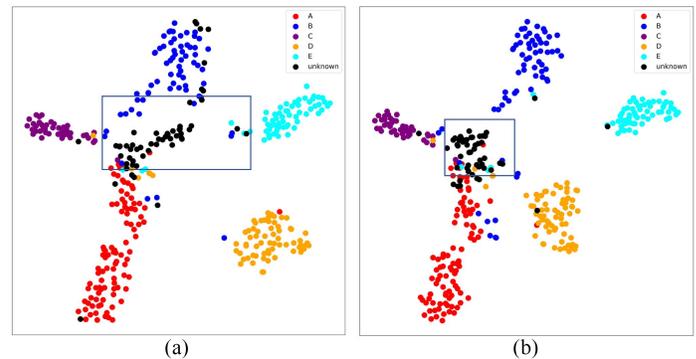

(a)　　　　　　　　　(b)

Fig. 8 Visualization of latent features based on t-SNE. (a) Feature space constructed by the baseline. (b) Feature space constructed by OSAD. A, B, C, D, E represent known categories (colored dots), while F, G, H represent unknown categories (black dots). Our proposed approach results in a more compact arrangement of known categories, reducing the overlap between known and unknown categories.

*F. Cross-validation*

To further validate the algorithm's generalizability, this work has conducted validation tests on the Zhong Ke Xing Tu (ZKXT) public dataset[7]. This dataset originates from the SAR aircraft target detection task of the 5th High-Resolution Remote Sensing Image Interpretation Software Competition,

# > REPLACE THIS LINE WITH YOUR MANUSCRIPT ID NUMBER (DOUBLE-CLICK HERE TO EDIT) <
13containing 2000 images, and image sizes range from 600 to 2048 pixels. These data undergo expert review and annotation, encompassing a total of 6556 annotated aircraft targets across seven distinct categories, including A320/321, Boeing787, A220, other, Boeing737-800, ARJ21, and A330. The download address of this data set can be found in https://github.com/xiayang-xiao/Dataset-zkxt. This work maintains consistent experimental settings with seven distinct tasks, where four tasks are associated with known categories (A320/321, Boeing787, A220, other), while keeping the rest of the experimental conditions consistent.

TABLE V presents the quantitative analysis results on the ZKXT dataset. The results indicate that the proposed method achieves detection of unknown categories while maintaining closed-set accuracy. Specifically, in task T-7, the proposed method achieves an average precision of 7.93% for detecting unknown categories. However, compared to the GF3 dataset, we observe a slightly lower average precision in detecting unknown categories. That could be speculated due to issues with the imaging quality of the images in the ZKXT dataset. The target features on the images in this dataset may not be distinct, leading to blurry imaging.

TABLE V Experimental results on the ZKXT data set

| Methods | T-1 | T-2 | | | | T-3 | | | | T-4 | | | |
|---|---|---|---|---|---|---|---|---|---|---|---|---|---|
| | $mAP_k\uparrow$ | $WI\downarrow$ | $AOSE\downarrow$ | $mAP_k\uparrow$ | $AP_u\uparrow$ | $WI\downarrow$ | $AOSE\downarrow$ | $mAP_k\uparrow$ | $AP_u\uparrow$ | $WI\downarrow$ | $AOSE\downarrow$ | $mAP_k\uparrow$ | $AP_u\uparrow$ |
| FR-CNN | 92.12 | 0 | 8 | 92.60 | 0 | 0 | 8 | 92.72 | 0 | 0.08 | 7 | 92.94 | 0 |
| OSAD | 92.83 | 0 | 1 | 92.91 | 2.02 | 0 | 5 | 93.04 | 0.19 | 0.08 | 3 | 93.16 | 0.27 |
| | | T-5 | | | | T-6 | | | | T-7 | | | |
| FR-CNN | -- | 0 | 18 | 92.51 | 0 | 4 | 23 | 92.87 | 0 | 3.84 | 34 | 92.23 | 0 |
| OSAD | -- | 0.95 | 13 | 92.76 | 5.23 | 1 | 14 | 92.92 | 7.65 | 2.61 | 26 | 92.47 | 7.93 |

*G. Sensitivity Analysis on Hyperparameters*

(1) The impact of stage in GCM

In Section III-C, this work describes the utilization of GCM with a bottleneck ratio of $r = 4$ applied to the feature layers P3, P4, P5. In this section, this work examines the impact of incorporating GCM at various stages. We observe these effects by incrementally introducing GC modules at different stages.

TABLE VI The effects of different stage

| | $WI\downarrow$ | $AOSE\downarrow$ | $mAP_k\uparrow$ | $AP_u\uparrow$ |
|---|---|---|---|---|
| Baseline | 4.48 | 55.00 | 95.47 | 0 |
| P3 | 4.41 | 42.00 | 95.40 | 9.23 |
| P4 | 4.46 | 46.00 | 95.80 | 9.86 |
| P5 | 4.42 | 37.00 | 96.21 | 10.13 |
| P3-P5 | 4.40 | 32.00 | 96.39 | 10.28 |

TABLE VI displays the performance results of the model following the integration of GCM at different stages. Notably, all stages exhibit improvements in performance attributed to the global context information modeling within GCM, underscoring the pivotal role of global context information in enhancing model performance. When GCM is introduced into c4 and c5, the model demonstrates superior performance compared to its insertion into c3, implying that higher-level semantic feature layers can more comprehensively leverage global context modeling. This phenomenon likely arises from the increment in model parameters as GCM is applied to multiple layers, thereby offering enhanced modeling capabilities and further augmenting performance. Furthermore, distinctions in the impact of GCM at different hierarchical levels may also contribute to these observations. For instance, integrating GCM at lower-level feature layers (e.g., P3) assists the model in capturing fine-grained details, including low-level textures and structural information. Conversely, introducing GCM at higher-level feature layers (e.g., P4 and P5) provides richer semantic insights, aiding in the abstraction of object representations and comprehension of higher-level semantics. In summary, the effective integration of the GCM substantiates its vital role in advancing model performance, with its insertion at various feature levels offering adaptability to diverse task requirements. This outcome further underscores the indispensable nature of global context information in the context of object detection tasks.

(2) Hyperparameters in LPG

In Section III-D, this work elucidates how LPG optimizes the traditional RPN network. The pertinent hyperparameters for LPG primarily encompass the number of proposal boxes ($Pro_{num}$), the threshold used to define positive and negative sample boxes ($Pos_{iou}$), and the proportion used to calculate the positive sample loss ($Pos_{ratio}$). The default settings in the RPN stage include $Pro_{num} = 256$, $Pos_{iou} = 0.7$, and $Pos_{ratio}=0.5$. We investigate alternative choices on the T-6 task that differ from the default settings, mainly focusing on $Pro_{num}$, $Pos_{iou}$ and $Pos_{ratio}$. The parameters utilized in the experiments are highlighted in black in the following table. For precise parameter values, please consult the TABLE VII.

TABLE VII Ablation study of Hyperparameters in LPG

| $Pro_{num}$ | 128 | 128 | 128 | 256 | **256** | 256 |
|---|---|---|---|---|---|---|
| $Pos_{iou}$ | 0.5 | 0.7 | 0.7 | 0.7 | **0.7** | 0.5 |
| $Pos_{ratio}$ | 0.5 | 0.5 | 0.7 | 0.7 | **0.5** | 0.5 |
| $WI\downarrow$ | 6.31 | 6.23 | 6.17 | 7.45 | **7.86** | 7.84 |
| $AOSE\downarrow$ | 45.00 | 43.00 | 48.00 | 78.00 | **57.00** | 82.00 |
| $mAP_k\uparrow$ | 95.32 | 96.10 | 95.89 | 94.12 | **95.55** | 95.36 |
| $AP_u\uparrow$ | 12.47 | 13.35 | 13.86 | 17.63 | **16.56** | 15.24 |

(3) The Queue Size in PCL

In Section III-E, this work provides a detailed explanation of how class-specific queues, denoted as qi, are employed to store feature vectors for computing category prototypes. Here, this work chooses to investigate the impact of the [26] hyperparameter Q on the learning task T-6 and document the experimental results in TABLE VIII. We observe that the performance remains relatively consistent across different Q values. This observation can be attributed to the fact that once prototypes are defined, they are periodically updated using newly observed features. Consequently, the specific quantity of features used to compute these prototypes is not critical. The choice of the Q value does not significantly influence performance because prototypes adapt and self-update based on newly observed features. For the sake of consistency, this work selects Q = 16 for all experiments.

TABLE VIII The impact Queue Size in PCL

| Q | WI↓ | AOSE↓ | $mAP_k$↑ | $AP_u$↑ |
|---|---|---|---|---|
| 6 | 6.98 | 43.00 | 95.01 | 15.46 |
| 8 | 6.34 | 45.00 | 95.42 | 15.93 |
| 12 | 7.92 | 50.00 | 94.93 | 17.02 |
| 16 | 7.86 | 57.00 | 95.55 | 16.56 |

*H. Failure Case Analysis*

One potential scenario where errors can easily occur is in the case of densely packed targets with significant differences in size. For instance, as illustrated in Fig. 9, objects labeled as "E" belong to small aircraft, densely clustered together, while object "H" is significantly larger but goes undetected during the detection process. It is worth noting that even the baseline algorithm failed to correctly detect "H," which may indicate shortcomings in the algorithm's ability to detect multi-scale objects.

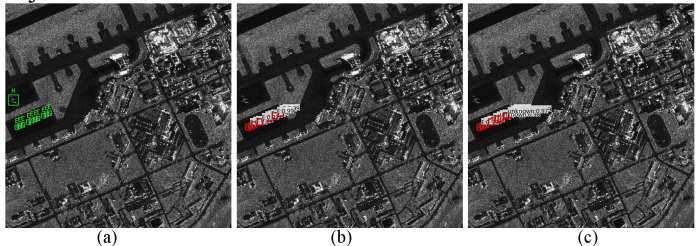

(a)        (b)        (c)

Fig. 9 (a), (b), and (c) represent the ground truth, baseline, and OSAD results, respectively. In (c), among a row of densely arranged small aircraft, some targets are identified as unknown instances, but not all. Additionally, the relatively larger aircraft "H" is missed in detection.

Our primary objective is currently to identify aircraft that do not belong to known categories, and the oil drum clearly falls outside the scope of our target of interest. As shown in Fig. 10, the proposed method successfully detects this oil drum and classify it as "unknown." However, from another perspective, the oil drum belongs to the newly discovered category of unknown objects.

In certain instances involving blurry images, the proposed model exhibits a tendency to make recognition errors. As shown in Fig.11, the detector missed an unknown class F and misclassified a known class C as an unknown class. This clearly highlights the limitations of our neural network when handling blurry images, whereas traditional closed-set methods perform better in this regard. This failure case not only underscores the challenges associated with processing blurry images but also serves as a reminder of the need for further model enhancements to improve recognition capabilities in such scenarios.

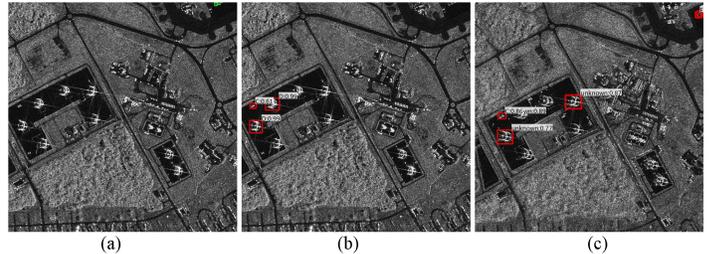

(a)        (b)        (c)

Fig. 10 (a), (b), and (c) represent the ground truth, baseline, and OSAD results, respectively. In (c), the oil drum is identified as an unknown category. Our objective is to identify unknown aircraft targets as unknown objects. The oil drum belongs to the newly discovered category of unknown objects.

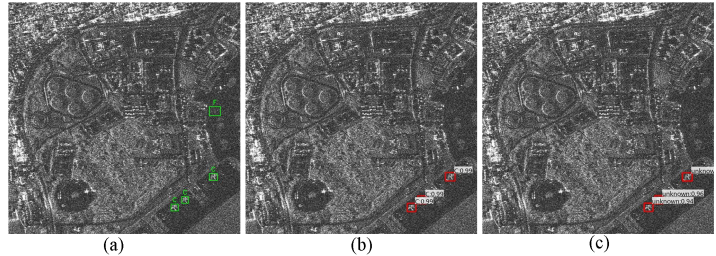

(a)        (b)        (c)

Fig. 11 (a), (b), and (c) represent the ground truth, baseline, and OSAD results, respectively. In (c), known category C is misclassified as an unknown category, and another unknown category F is missed.

V. CONCLUSION

This paper proposes an innovative solution called OSAD, designed to address the complexity of SAR image open-set detection tasks. The OSAD approach breaks away from the CNN paradigm operating under the closed-world assumption. Firstly, to comprehensively capture global contextual semantic information in SAR images, this work introduces GCM following the original backbone network. This augmentation enhances the model's feature extraction capabilities, particularly for the target of interest. Secondly, this work replaces the traditional RPN with LPG to bolster the model's generalization capacity towards unknown objects. This modification enables more comprehensive pseudo-label generation for unknown objects. Finally, a discriminative scheme based on prototype distances is designed to achieve superior discriminative performance for unknown categories. The organic integration of these key components empowers OSAD to successfully accomplish open-set detection tasks for aircraft targets in SAR images.

However, during the analysis of failure cases, this study encounters several challenges, especially when dealing with blurry images. In some instances, proposals for known categories are erroneously classified as unknown objects during testing, and conventional non-maximum suppression (NMS) filtering failed to rectify these misclassifications. This may be attributed to the distant feature distribution between certain known categories and category centers, resulting in these misclassifications. Consequently, one of our prospective research directions involves the exploration of methods to effectively constrain the distance between known category samples and category centers, thereby mitigating the occurrence of such misclassifications.